\documentclass[runningheads]{llncs}
\usepackage{cite}
\usepackage{xspace}
\usepackage{tikz,pifont}
\usepackage{amsmath,amssymb,amsfonts}
\usepackage{listings}
\usepackage{graphicx}
\usepackage{textcomp}
\usepackage{xcolor, colortbl}
\usepackage{makecell}
\usepackage{hyperref}
\usepackage{xcolor}
\usepackage{amssymb}
\usepackage[many]{tcolorbox}
\usepackage{subcaption}

\usepackage{enumitem}
\newlist{questions}{enumerate}{2}
\setlist[questions,1]{label=\textbf{RQ\arabic*.},ref=\textbf{RQ\arabic*}}
\setlist[questions,2]{label=(\alph*),ref=\thequestionsi(\alph*)}

\usepackage[ruled, vlined]{algorithm2e}
\SetKwInput{KwInput}{Input}
\SetKwInput{KwOutput}{Output}
\usepackage{booktabs}

\newcommand*{\ie}{i.e.,\@\xspace}
\newcommand*{\eg}{e.g.,\@\xspace}

\makeatother
\newcommand*{\etal}{\emph{et~al.}\@\xspace}

\definecolor{verylightgray}{gray}{0.92}
\definecolor{ao(english)}{rgb}{0.0, 0.5, 0.0}

\definecolor{deepblue}{rgb}{0,0,0.5}
\definecolor{deepred}{rgb}{0.6,0,0}
\definecolor{deepgreen}{rgb}{0,0.5,0}
\definecolor{shadecolor}{gray}{0.9}

\colorlet{shadecolor}{verylightgray}
\colorlet{framecolor}{black}

\newtcolorbox{shadedbox}{
drop shadow southeast,
breakable,
enhanced jigsaw,
colback=white,
boxrule=0.80pt,
left=0.3em,
right=0.3em,
top=0.1em,
bottom=0.05em
}

\definecolor{mygreen}{rgb}{0,0.6,0}
\definecolor{mygray}{rgb}{0.95,0.95,0.95}
\definecolor{myred}{rgb}{0.5,0,0}

\definecolor{verylightgray}{gray}{0.92}
\definecolor{ao(english)}{rgb}{0.0, 0.5, 0.0}

\newcommand{\codelink}{\url{https://shorturl.at/DGMX1}}

\newcommand{\SoBigDataITAck}{European Union - NextGenerationEU - National Recovery and Resilience Plan (Piano Nazionale di Ripresa e Resilienza, PNRR) - Project: “SoBigData.it - Strengthening the Italian RI for Social Mining and Big Data Analytics” - Prot. IR0000013 - Avviso n. 3264 del 28/12/2021\xspace}

\newcommand{\HubAck}{"ICSC – Centro Nazionale di Ricerca in High Performance Computing, Big Data and Quantum Computing", funded by European Union – NextGenerationEU\xspace}

\begin{document}
\title{Towards a Prediction of Machine Learning Training Time to Support Continuous Learning Systems Development. \thanks{This work is partially supported by \HubAck, by "Data-quality-driven estimation of computational complexity of Machine Learning systems" project, funded by University of L'Aquila, 2023, and by \SoBigDataITAck}
}
\titlerunning{Towards a Prediction of Machine Learning Training Time}
%
\author{Francesca Marzi\orcidID{0009-0009-9129-9231} \and
Giordano d'Aloisio\orcidID{0000-0001-7388-890X} \and
Antinisca Di Marco\orcidID{0000-0001-7214-9945} \and Giovanni Stilo\orcidID{0000-0002-2092-0213}}
\authorrunning{F. Marzi et al.}

\institute{University of L'Aquila, L'Aquila, Italy\\
\email{
giordano.daloisio@graduate.univaq.it\\
\{francesca.marzi,antinisca.dimarco,giovanni.stilo\}@univaq.it}}
\maketitle              
\begin{abstract}

The problem of predicting the training time of machine learning (ML) models has become extremely relevant in the scientific community. Being able to predict \textit{a priori} the training time of an ML model would enable the automatic selection of the best model both in terms of energy efficiency and in terms of performance in the context of, for instance, MLOps architectures.
In this paper, we present the work we are conducting towards this direction. In particular, we present an extensive empirical study of the Full Parameter Time Complexity (FPTC) approach by Zheng \etal, which is, to the best of our knowledge, the only approach formalizing the training time of ML models as a function of both dataset's and model's parameters. We study the formulations proposed for the Logistic Regression and Random Forest classifiers, and we highlight the main strengths and weaknesses of the approach. Finally, we observe how, from the conducted study, the prediction of training time is strictly related to the context (\ie the involved dataset) and how the FPTC approach is not generalizable. 

\end{abstract}

\section{Introduction}

The problem of energy efficiency and sustainability of machine learning (ML) systems is becoming increasingly important within the scientific community \cite{fischer2022unified,wenninger2022sustainable,garcia2019estimation}, as also highlighted by the ONU's Sustainable Development Goals (\eg Goal 9 or Goal 12)\cite{onu_sdg}. Generally, the energy consumption of ML models is directly related to the \textit{training phase time complexity}. This means that the longer it takes to train a model, the more energy is required by the system. For this reason, predicting \textit{a priori} the training time of an ML model will be a significant advance in such direction, enabling the automatic selection of the efficient ML model. 
The training time prediction of ML models also becomes highly relevant in the context of MLOps and, in general, \textit{continuous learning} or \textit{learning-enabled} systems, where the ML model is constantly re-trained with new data \cite{alla2021mlops}. As highlighted in \cite{nahar_meta-summary_2023}, engineering such kind of system is always very challenging since the development processes are often ad-hoc and specific to the use case. For this reason, having an \textit{a priori} estimation of the training time can help in standardizing some phases of the development process in contexts where, for instance, the computational power for training the model is very limited (\eg, IoT devices \cite{zikria2020deep}). In addition, selecting the most efficient ML model can help stakeholders satisfy other relevant quality properties of software architectures, like \textit{performance} \cite{lewis2021software}.

In this paper, we present the work we are conducting towards a prediction of ML training time. In particular, we present an extensive empirical evaluation of the Full Parameter Time Complexity (FPTC) approach proposed by Zheng \etal in \cite{zheng2021full}, which is, to the best of our knowledge, the only approach so far that formulates the ML training time as a function of dataset's and ML model's parameters. Specifically, differently from what has been done in \cite{zheng2021full}, where the authors use only one dataset, we use the FPTC approach to predict the training time of a Logistic Regression \cite{menard_applied_2002} and Random Forest \cite{rigatti2017random} classifier on a heterogeneous set of data, and we compare the predicted time with the actual training time of the method, highlighting the main strengths and weaknesses of the approach\footnote{The replication package of the experiments is available here: \codelink}.

The paper is structured as follows: in Section \ref{sec:related} we discuss some related works in the context of training time prediction; Section \ref{sec:background} describes in detail the FPTC approach; Section \ref{sec:experiment} presents the conducted experiment and the research questions we want to answer; Section \ref{sec:results} shows the experiment's results and discuss them w.r.t. the research questions; finally Section \ref{sec:conclusion} presents some future works and concludes the paper.

\vspace{-.2cm}
\section{Related Work}\label{sec:related}
\vspace{-.3cm}
Nowadays, the estimation of the running time of the training phase of ML models is primarily conducted through empirical analysis relying on a set of common characteristics.

In \cite{kwon2013effects}, the authors performed empirical analyses to assess the impact of different dataset characteristics, such as sample size, class type, missing values and dimensionality, on the performance of classification algorithms, considering both accuracy and elapsed time.
In \cite{ali2006learning}, a rule-based learning algorithm was derived through an empirical evaluation of the performance of eight classifiers on 100 classification datasets, comparing them based on various accuracy and computational time measures.
The empirical results were combined with the dataset characteristic measures to formulate rules to determine which algorithms were best suited for solving specific classification problems. 
Finally, in \cite{mohr2021predicting}, a model was developed to predict the running time of ML pipelines through empirical analysis of different ML algorithms with a heterogeneous set of data. The approach was used to predict the timeout of an ML pipeline. 

Considering non-empirical analyses, to the best of our knowledge, \cite{zheng2021full} is the first attempt to provide an a priori estimation of the training time for various ML models without actually running the code. 
In this work, the authors propose a method to quantitatively evaluate the time efficiency of an ML classifier called Full Parameter Time Complexity (FPTC). The authors derive FPTC for five classification models, namely Logistic Regression, Support Vector Machine, Random Forest, K-Nearest Neighbors, and Classification and Regression Trees. FPTC depends on several variables, including the number of attributes, the size of the training set, and intrinsic characteristics of the algorithms, such as the number of iterations in Logistic Regression or the number of Decision Trees in a Random Forest. 
A coefficient $\omega$ was introduced to establish the relationship between the running time and FPTC. The coefficient $\omega$ can be obtained through a preliminary experiment on a small sampled dataset under different execution environments. When the physical execution environment changes, the coefficient $\omega$ should be reevaluated to reflect the new conditions.

Based on this state-of-the-art analysis, we observe that most of the studies concerning the training time of ML models tend to rely on empirical approaches. The only approach formalizing the training time as a function of datasets' and ML models' parameters is \cite{zheng2021full}. In this paper, we aim to highlight the strengths and weaknesses of this approach by conducting an extensive evaluation of the method. 


\section{Background Knowledge}\label{sec:background}

In this section, we describe in detail the FPTC method \cite{zheng2021full} where the training time of several ML models is defined as a function of different parameters of the dataset, of the model itself, and of a coefficient ($\omega$)  that reflects the influence given by the execution environment on the actual training time of the model. This value should vary only when an ML model runs on a different execution environment.  We detail better in Section \ref{sec:experiment} how $\omega$ has been computed in our experiment. In this work, we focus on the formulation of the training time for two particular ML models, \ie Logistic Regression (\textit{LogReg}) \cite{menard_applied_2002} and Random Forest (\textit{RF}) \cite{rigatti2017random}, while we leave the analysis of other methods to future works. 

The FPTC for the Logistic Regression classifier is defined as:
\begin{equation}\label{eq:fptc_logreg}
    FPTC_{LogReg} = F(Qm^2vn) * \omega_{LogReg}
\end{equation}
where $n$ is the number of rows of the dataset, $v$ is the number of dataset's features, $m$ is the number of classes of the dataset, $Q$ is the number of model's iterations during the training phase, and $\omega_{LogReg}$ is the slope of a regression function computed comparing the results of the first part of the equation \ref{eq:fptc_logreg} with the actual training time of a Logistic Regression model using a subset of the training datasets.

The FPTC for the Random Forest classifier is defined instead as:
\begin{equation}
    \label{eq:fptc_rf}
    FPTC_{RF} = F(s(m+1)nv\log_{2}(n)) * \omega_{RF}
\end{equation}
where $n$, $m$, and $v$ are the same variables as above, while $s$ is the number of trees of the random forest. $\omega_{RF}$ is again defined as the slope of a regression function computed comparing the results of the first part of the equation \ref{eq:fptc_rf} with the actual training time of a Random Forest classifier on a set of synthetic datasets. 

Concerning $\omega$, the authors state that this variable reflects the influence given by the execution environment on the actual training time of the model. Hence, this value should vary only when an ML model runs on a different environment. We detail better in Section \ref{sec:experiment} how $\omega$ has been computed in our experiment.  
\section{Experimental Setting}\label{sec:experiment}

This section describes the experiments we conducted to evaluate the FPTC method. In particular, with our experiments, we aim to answer the following two research questions:

\begin{questions}
    \item\label{rq1} Is the slope ($\omega$) parameter of FPTC only dependent on the execution environment?
    \item\label{rq2} Is the FPTC able to predict the training time of an ML model?
\end{questions}

In Section \ref{sec:slope_computation}, we describe the experimental setting conducted to compute the slope parameter. While in Section \ref{sec:training_time}, we describe the experiment led to predict the training time of the Logistic Regression and Random Forest models.
All the experiments have been executed on a DELL XPS 13 2019 with a processor Intel Core i7, 16GB of RAM and Ubuntu 22.04.2 LTS. 
\vspace{-.3cm}
\subsection{Slope Computation}\label{sec:slope_computation}

To answer \ref{rq1}, we must assess if the slope computation only depends on the execution environment. That is, given the same environment and the same ML model, the slope should not change significantly if the dataset used to compute the slope changes. To answer this question, we performed an experiment that computes a set of slopes using a synthetic dataset $D_s$  with 6,167 rows and 10,000 features. In particular, we calculate a set of slopes corresponding to 19 subsets of $D_s$, each one with a different subset of features. Next, we compared the different slopes obtained. It is worth noticing that, in \cite{zheng2021full}, the authors compute the slope on the same dataset on which they want to predict the training time. In this experiment, we use a synthetic dataset different from the ones on which we predict the training time. We have chosen a synthetic dataset instead of a real one to have better control over its number of features and instances. In addition, a synthetic dataset can be easily released and used for computing the slopes in further experiments.

\begin{algorithm}[ht!]
\caption{Slope computation}
\label{alg:slope}
\KwInput{(Synthetic dataset $D_s$, ML Model $M$, Number of starting features $f = 501$, Number of features to add $a=501$, Number of starting rows $s=100$, Number of rows to add $p=1,000$)}
\KwOutput{(List of slopes at increasing number of features)}
    $n = $ number of rows of $D_s$ \tcp{in our case 6.167}
    $m' = $ number of features of $D_s$ \tcp{in our case 10.000}
    $slopes = \{\}$\\
    \For{$i \in 20$}{
        $D_s' = $ subset of $D_s$ with $f$ features\\
        \While{features of $D_s' < m'$}{
            \textit{tt} = []\\
            \textit{fptcs} = []\\
            $m$ = features of $D_s'$\\
            \tcc{split D' into sub-datasets and get training times and fptc}
            \For{($r=s ; r<n ; r+=p$)}{
                 $D_s'' = $ dataset of $r$ rows from $D_s'$\\
                train $M$ on $D_s''$\\
                \textit{t} = training time of $M$\\
                \textit{fptc} = \textit{getFPTC($D_s''$, $M$)}\\
                add $t$ to \textit{tt}\\
                add \textit{fptc} to \textit{fptcs}
            }
            \textit{reg} = \textit{LinearRegression()}\\
            train \textit{reg} on \textit{tt} and \textit{fptcs}\\
            $\omega = $ slope of \textit{reg}\\
            append $\omega$ to $slopes[m]$\\
            $D_s' = D_s' + a$ other features from $D_s$
        }
    }
    \For{$m \in slopes$ keys}{
        $slopes[m] =$ median of $slopes[m]$
    }
    \Return{slopes}
\end{algorithm}

Algorithm \ref{alg:slope} shows the procedure we followed to compute the slopes. The algorithm takes as input a synthetic dataset $D_s$, an ML model $M$ (in our case, $M$ is either a Logistic Regression or a Random Forest classifier), and a set of parameters useful for the analysis: $f$, \ie the number of starting features of the synthetic dataset $D_s$; $a$, \ie the number of features to add at each iteration; $s$, \ie the number of rows of the first sub-dataset used to compute the slope; and $p$, \ie the number of rows to add to each other sub-dataset. In our case, $f=501$, $a=501$, $s=100$, and $p=1.000$. The algorithm returns a list of slopes, each one corresponding to a subset $D_s'$ of $D_s$ with a number of features lower or equal to the ones in $D_s$. At the first iteration, $D_s'$ has 501 features. Next, $D_s'$ is split into a set of sub-datasets $D_s''$ with an increasing number of rows ranging from 100 to the total number of rows. Each sub-dataset has a delta of 1000 rows. These sub-datasets are used to compute the training time of the model $M$ and the relative \textit{FPTC} prediction using equations \ref{eq:fptc_logreg} and \ref{eq:fptc_rf} for Logistic Regression and Random Forest, respectively. After computing the training times and the \textit{FPTC} predictions for each sub-dataset $D_s''$, the training times and the \textit{FPTC} predictions are used to train a \textit{Linear Regression} model and to get its slope $\omega$. The obtained slope is added to a dictionary of slopes with the key equal to the number of features of $D_s'$. Finally, the number of features of $D_s'$ is increased by 500. This procedure continues until the number of features of $D_s'$ equals the number of features of $D_s$. This whole process is repeated 20 times, and the median slope of each subset $D_s'$ is finally returned.

\subsection{Training Time Prediction}\label{sec:training_time}

To answer the \ref{rq2}, we conducted a set of experiments to predict, using the FPTC method, the training time of a Logistic Regression and a Random Forest classifier using 7 heterogeneous datasets. Then we compared the predicted training time with the actual training time of the method.
\vspace{-.4cm}
\begin{algorithm}[ht!]
    \caption{Training time prediction}
    \label{alg:prediction}
    \KwInput{(Dataset $D$, ML Model $M$, List of slopes $S$)}
    \KwOutput{(List of Root Mean Squared Errors $RMSE$, List of Mean Absolute Percentage Error $MAPE$)}
        $trainingTimes = []$\\
        \For{$i \in 100$}{
            train $M$ on $D$\\
            $t =$ training time of $M$\\
            add $t$ to $trainingTimes$\\
        }
        $tt = mean(trainingTimes)$\\
        $RMSE = []$\\
        $MAPE = []$\\
        \For{$\omega \in S$}{
            $FPTC = $\textit{getFPTC($D$, $M$, $\omega$)}\\
            $rmse = getRMSE(tt, FPTC)$\\
            $mape = getMAPE(tt, FPTC)$\\
            add $rmse$ to $RMSE$\\
            add $mape$ to $MAPE$\\
        }
        \Return{RMSE, MAPE}
\end{algorithm}
\vspace{-.4cm}
Algorithm \ref{alg:prediction} reports the experiment's pseudo-code. The algorithm takes as input a dataset $D$, the ML model $M$, and the list of slopes $S$ computed with the procedure described in Algorithm \ref{alg:slope}, and returns a list of Root Mean Squared Errors $RMSE$ \cite{chai2014root} and Mean Absolute Percentage Errors $MAPE$ \cite{de2016mean}, one for each slope. The experiment can be divided into two steps. In the first step, the algorithm computes 100 times the training time of the ML model $M$ on $D$ and then calculates the mean of the times. In the second step, for each slope, $\omega$, the algorithm computes the \textit{FPTC} and the RMSE and MAPE between the actual training time and the \textit{FPTC}. Finally, the list of errors is returned.

In the evaluation, we have employed 7 heterogeneous datasets which differ in terms of dimensions to evaluate if the FPTC method works better under datasets. The list of employed datasets is reported in Table \ref{tab:values}\footnote{
Before running Algorithm \ref{alg:prediction}, following the guidelines reported in \cite{scikit-learn}, all the data has been scaled by removing the mean ($\mu$) and by dividing the variance ($\sigma$) from each feature.  
}.

Concerning the ML classifiers, we used the implementations from the \textit{scikit-learn} library \cite{scikit-learn} and, following the hyper-parameters settings of \cite{zheng2021full}, we set the \textit{l2} penalty and \textit{sag} solver for the Logistic Regression, while we set the number of trees of the Random Forest classifier to 80. Finally, we set the maximum number of iterations of the Logistic Regression to 10.000.
\vspace{-.5cm}
\begin{table}[ht!]
\centering
\caption{Values of FPTC parameters for each dataset}
\label{tab:values}
\begin{tabular}{|l|c|c|c|c|c|}
\hline
&  \multicolumn{3}{c|}{\textbf{Dataset Coefficients}}  &  \multicolumn{2}{c|}{\textbf{ML Methods Coefficients}} \\ \hline
\textbf{Dataset} & \textbf{Instances} & \textbf{Features} & \textbf{Classes} & \textbf{LogReg Iters} & \textbf{RF Trees}\\ \hline
Adult \cite{kohavi1996scaling} & 30940 & 101 & 2 & 635 & 100 \\ \hline
Antivirus \cite{misc_detect_malacious_executable(antivirus)_355} & 373 & 531 & 2 & 840 & 100 \\ \hline
APS \cite{misc_aps_failure_at_scania_trucks_421} & 60000 & 162 & 2 & 5068.73 & 100 \\ \hline
Arcene \cite{misc_arcene_167} & 100 & 10000 & 2 & 1089 & 100 \\ \hline
Compas \cite{angwin_machine_2016} & 6167 & 400 & 2 & 721 & 100 \\ \hline
Dexter \cite{misc_dexter_168} & 300 & 20000 & 2 & 855.91 & 100 \\ \hline
German \cite{ratanamahatana2002scaling} & 1000 & 59 & 2 & 33.93 & 100 \\ \hline
\end{tabular}
\end{table}

Table \ref{tab:values} synthesizes, for each dataset, the values of the different parameters of the two FPTC formulations for Logistic Regression and Random Forest classifiers. In particular, together with the dimensions of the datasets, we also report the number of iterations required by the Logistic Regression to train and the number of trees of the Random Forest.

\section{Experimental Results and Discussion}\label{sec:results}

In this section, we present the results of our experimental evaluation and discuss them with respect to the research questions defined in Section \ref{sec:experiment}. Finally, we present some threats to validity of our evaluation.

\subsection{Addressing \ref{rq1}}\label{sec:exp_one}

\begin{figure}[ht!]
    \centering
    \begin{subfigure}{\textwidth}
        \centering
        \includegraphics[width=.8\textwidth]{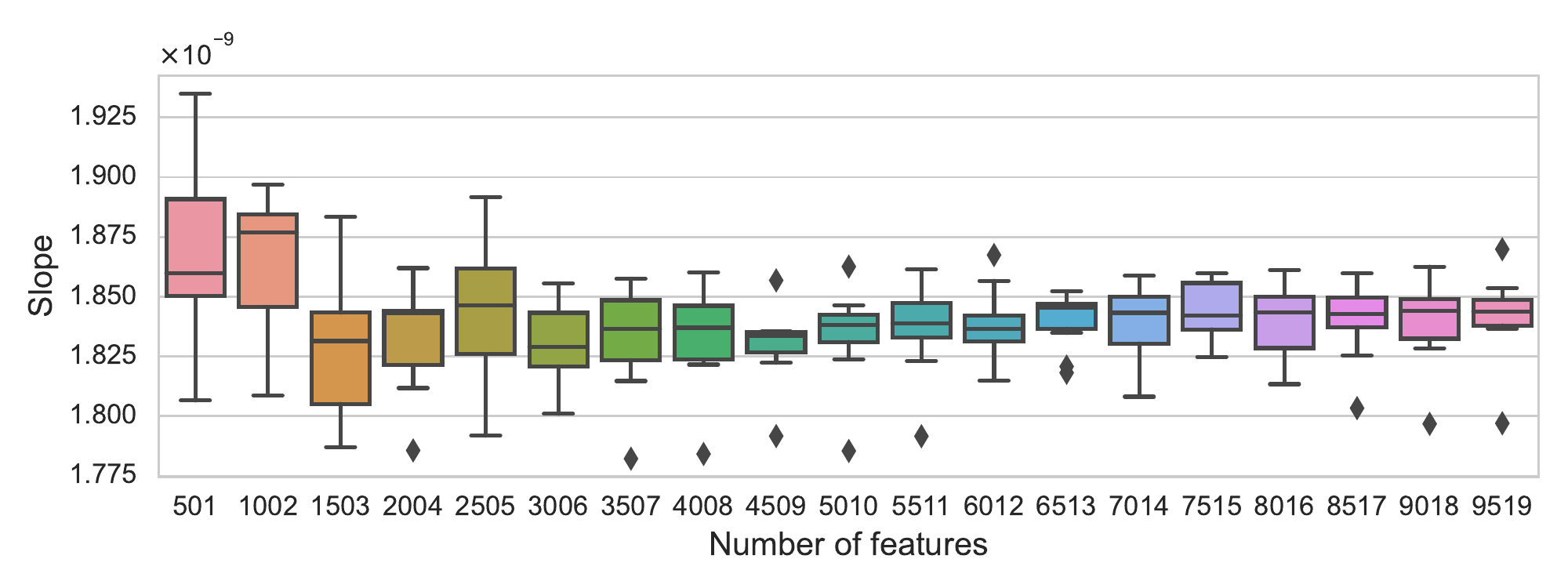}
        \caption{Logistic Regression}
        \label{fig:slope_var_logreg}
    \end{subfigure}
    \begin{subfigure}{\textwidth}
        \centering
        \includegraphics[width=.8\textwidth]{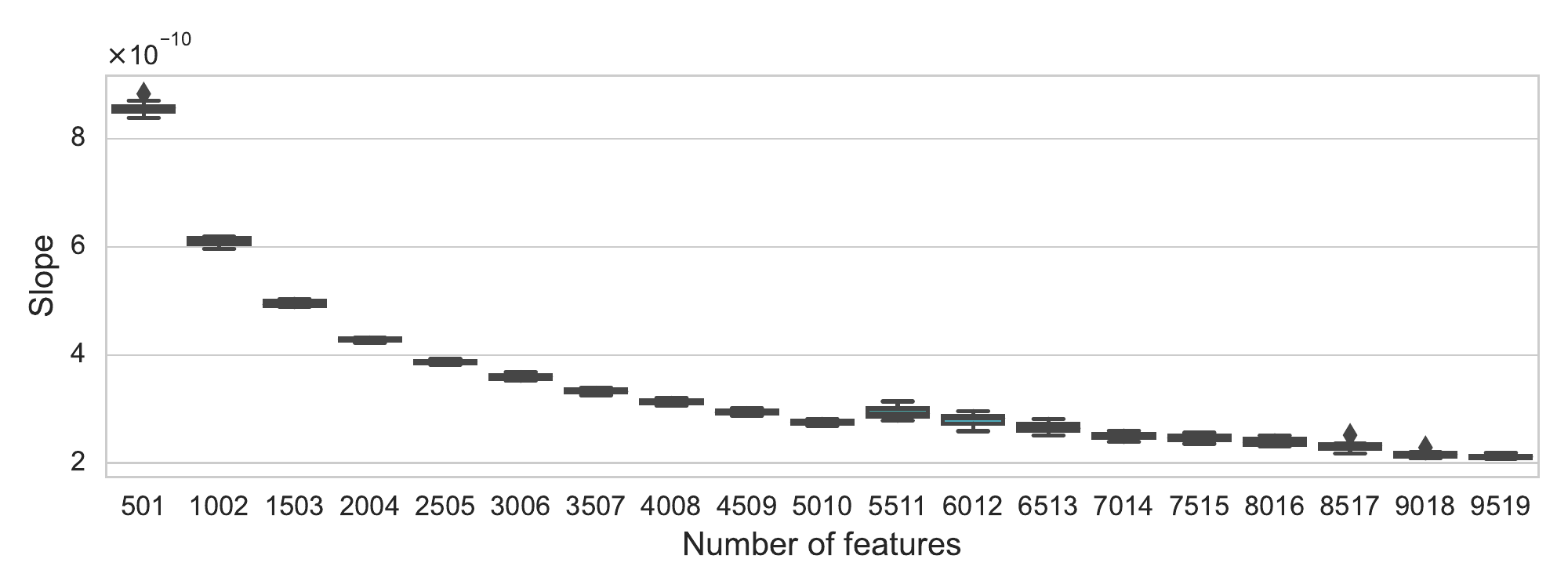}
        \caption{Random Forest}
    \label{fig:slope_var_rf}
    \end{subfigure}
    \caption{Slope variation with an increasing number of dataset's features}
    \label{fig:slope_var}
\end{figure}

Figure \ref{fig:slope_var} reports the boxplot of the variation of the slopes computed with an increasing number of features of the synthetic dataset. In particular, figure \ref{fig:slope_var_logreg} reports the slopes computed for the Logistic Regression classifier, while figure \ref{fig:slope_var_rf} reports the slopes computed for the Random Forest classifier.

Concerning the Logistic Regression model, it can be seen (in figure \ref{fig:slope_var_logreg}) how the slopes have generally low variability.
An exception is given by the slopes computed with 501 and 1002 features which are, on average, higher than the others. In particular, the median of the slopes computed using 501 features is around 0.02 points higher than the others, while the median of the slopes calculated using 1002 features is about 0.04 points higher than the others. In all the other cases, the median slope ranges from $1.83*10^{-9}$ to $1.85*10^{-9}$. 

Concerning the Random Forest classifier, it can be seen from figure \ref{fig:slope_var_rf} how the slopes present a higher variability among them, starting from a value around $8.5*10^{-10}$ using 501 features to a value of $2*10{-10}$ using 9519 features. In particular, it can be noticed from the figure that the value for the slope tends to decrease with an increase in the number of the dataset's features. 


Moreover, we study the significance of the results of the slopes by performing the ANOVA test \cite{mcdonald2009handbook} for both experiments. This test checks for the null hypothesis that all groups (\ie all the slopes computed using the same number of features) have the same mean; if the confidence value (\textit{p-value}) is $>0.05$, the null hypothesis is confirmed. Concerning the Logistic Regression classifier, the test returned a \textit{p-value} of 0.002, meaning the groups do not have the same mean. However, performing the same ANOVA test excluding the slopes computed with 501 and 1,002 features returns a \textit{p-value} of 0.352, accepting the null hypothesis of the same mean. This means that, excluding the slopes computed with 501 and 1.002 features, all the others have the overall same mean. Concerning the Random Forest classifier, the \textit{p-value} returned is $9.022*10^{-222}$, confirming the high variability of the slopes.

From this analysis of the slope variations, we can conclude how, differently from what is stated in \cite{zheng2021full}, the slopes do not change only when the execution environment changes, but they are also related to the number of features of the dataset used to compute them, in particular when using a Random Forest classifier.

\begin{shadedbox}
    \textbf{Answer to \ref{rq1}:}
    The slopes computed under the same execution environment but using an increasing number of features are pretty stable for the Logistic Regression classifier. Instead, they present a higher variance for the Random Forest classifier. Hence, we can conclude how the slope is also related to the number of features of the dataset used to compute them.
\end{shadedbox}

\subsection{Addressing \ref{rq2}}\label{sec:rq2}

\begin{figure}[ht!]
    \centering
    \includegraphics[width=.9\textwidth]{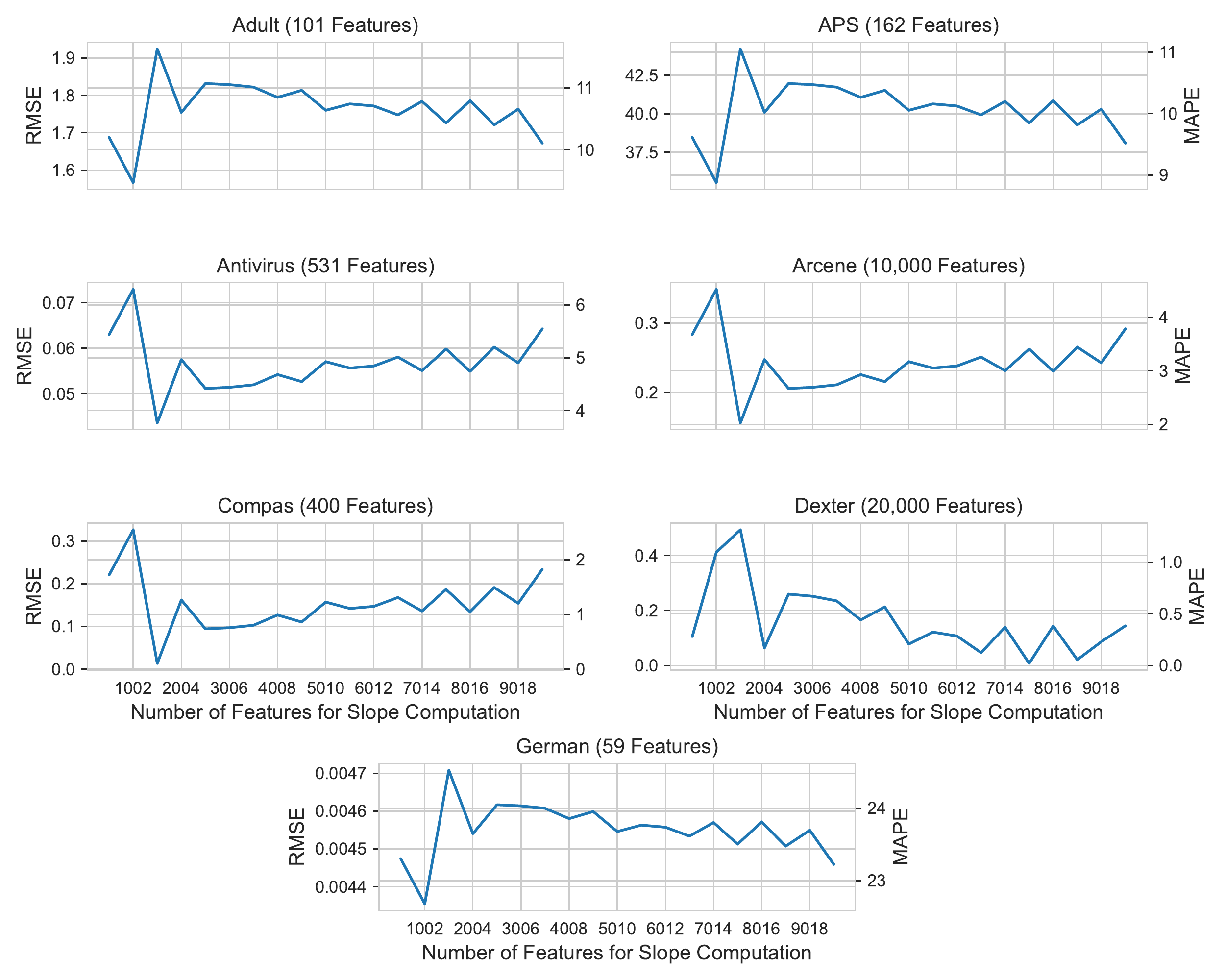}
    \caption{RMSE and MAPE at different slope values for LogReg}
    \label{fig:error_logreg}
\end{figure}

\begin{figure}[ht!]
    \centering
    \includegraphics[width=.9\textwidth]{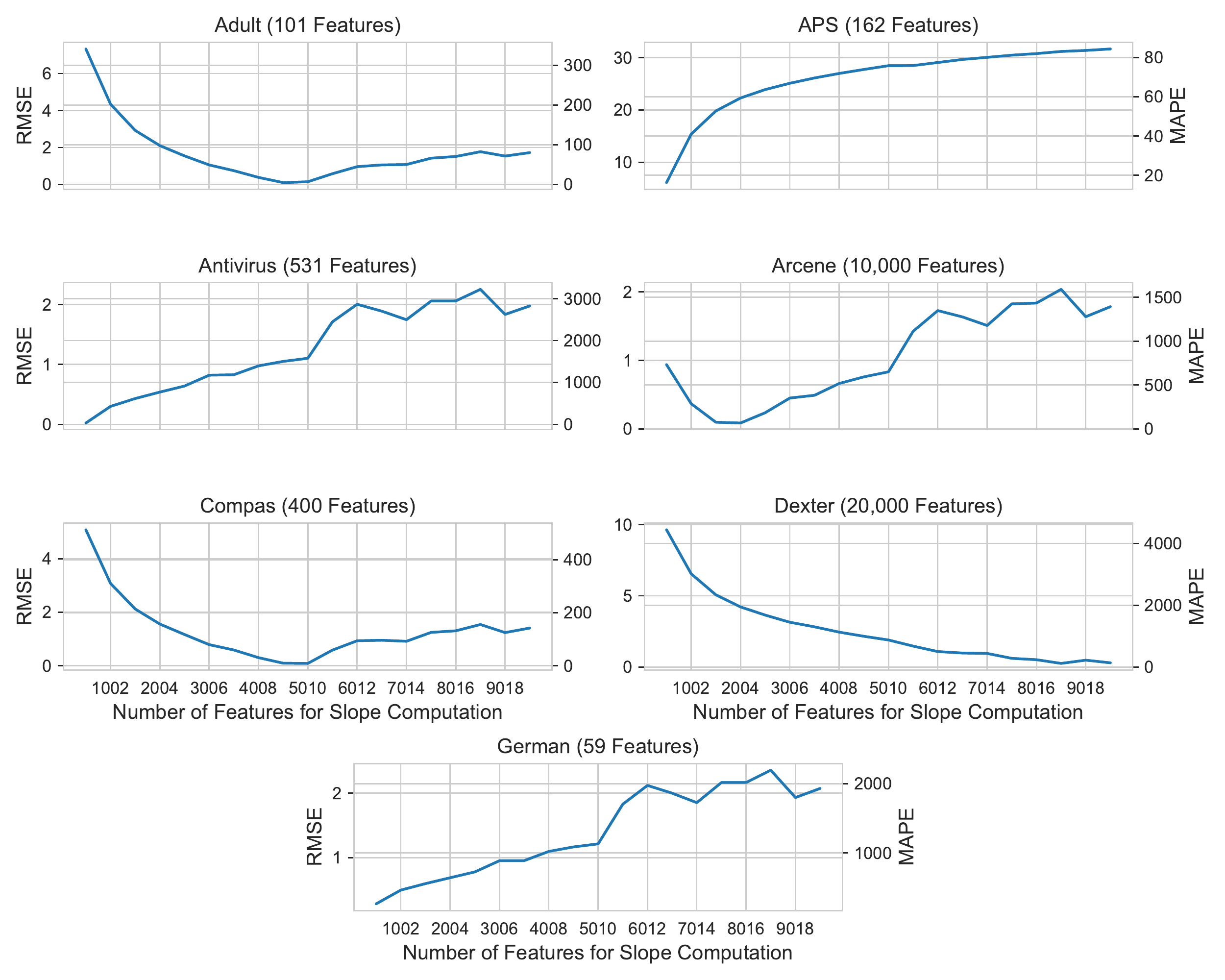}
    \caption{RMSE and MAPE at different slope values for Random Forest}
    \label{fig:errors_rf}
\end{figure}

Figures \ref{fig:error_logreg} and \ref{fig:errors_rf} report the errors in the predictions of the FPTC method compared to the actual training time of the Logistic Regression and Random Forest Classifier, respectively, for all the datasets described in Section \ref{sec:experiment}. In particular, in each figure, the left y-axis reports the RMSE, while the right y-axis reports the MAPE. On the x-axis, we report the number of features of the synthetic dataset used to compute the relative slope. Near each dataset name, we also report its number of features.

Concerning the Logistic Regression classifier, it can be seen from figure \ref{fig:error_logreg} how the FPTC method can predict the training time of the model under some datasets while it fails in the prediction of others. In particular, the FPTC method can predict the training time of the LogReg under the \textit{Antivirus} dataset (with an RMSE and MAPE almost equal to 0 using the slope computed with 9,009 features of the synthetic dataset), \textit{Arcene} (with an RMSE and MAPE almost equal to 0 using the slope computed with 6,006 features), \textit{Compas} (with an RMSE and MAPE almost equal to 0 using the slope computed with 4,004 features), and \textit{Dexter} (with an RMSE and MAPE almost equal to 0 using the slope computed with 501 features). In contrast, the FPTC method is not able to predict the training time of the LogReg under \textit{Adult} (with the lowest MAPE equal to 9.5 using the slope computed with 1,503 features), and \textit{APS} (with the lowest MAPE equal to 9.0 using the slope computed with 1,503 features).  It is worth noting that the high MAPE for the \textit{German} dataset may be influenced by the low values of FPTC and true running time, causing this metric to increase \cite{de2016mean}. This is also supported by a low value of the RMSE.   
\vspace{-.8cm}
\begin{table}[ht!]
    \centering
    \caption{Mean and stand. dev. of training time and FPTC for LogReg model}
    \label{tab:tt_fptc_logreg}
    \begin{tabular}{|l|c|c|}
    \hline
    \textbf{Dataset} & \textbf{Training Time (seconds)} & \textbf{FPTC (seconds)} \\
    \hline
     Adult & 16.54 $\pm$ 0.042 & 14.77 $\pm$ 0.066 \\
     \hline
     Antivirus & 1.15 $\pm$ 0.012 & 1.214 $\pm$ 0.006 \\
     \hline
     APS &  400.156 $\pm$ 1.126 & 356.81 $\pm$ 1.803 \\
     \hline
     Arcene & 7.711 $\pm$ 0.012  & 7.953 $\pm$ 0.006\\
     \hline
     Compas &  12.802 $\pm$ 5.366 & 12.956 $\pm$ 0.065 \\
     \hline
     Dexter &  37.597 $\pm$ 0.403 & 37.5 $\pm$ 0.188\\
     \hline
     German & 0.019 $\pm$ 0.003 & 0.015 $\pm$ $7.342 * 10^{-5}$ \\
     \hline
    \end{tabular}
\end{table}
\vspace{-.3cm}
Table \ref{tab:tt_fptc_logreg} reports the mean and standard deviation of the training time and FPTC in seconds for each selected dataset. From this table, it can be seen how the FPTC method tends to underestimate the real training time, especially in \textit{Adult} (with a delta of almost 2 seconds between the actual training time and the predicted one), and \textit{APS} (with a delta of almost 50 seconds between the actual training time and the predicted one). Finally, following the low variability of the slopes computed in Section \ref{sec:exp_one}, we notice how the slopes' variation does not much influence the FPTC predictions.

Figure \ref{fig:errors_rf} reports the same metrics computed for the Random Forest classifier. Differently from the Logistic Regression classier, here we notice how the FPTC method is more sensitive to the variation of the slopes, which lets the prediction increase or decrease significantly. This behaviour is explained by the high variability of the slopes shown in Section \ref{sec:exp_one}. In addition, it can be seen from the charts that the FPTC method can always predict real training time under a specific slope value achieving a value of zero for both RMSE and MAPE. However, we also notice how the value of the slope leading to the optimal predictions is not constant and varies between the datasets. The only dataset on which the FPTC method is not able to correctly predict the training time is the \textit{APS} dataset, with the lowest MAPE of around 15 points.
\vspace{-.8cm}
\begin{table}[ht!]
    \centering
    \caption{Mean and stand. dev. of training time and FPTC for RF model}
    \label{tab:tt_fptc_rf}
    \begin{tabular}{|l|c|c|}
    \hline
    \textbf{Dataset} & \textbf{Training Time (seconds)} & \textbf{FPTC (seconds)} \\
    \hline
     Adult & 2.15 $\pm$ 0.012 & 2.60 $\pm$ 2.383 \\
     \hline
     Antivirus & 0.07 $\pm$ $8.368 * 10^{-17}$  & 1.20 $\pm$ 0.711 \\
     \hline
     APS &  37.54 $\pm$ 0.698 & 11.49 $\pm$ 6.469 \\
     \hline
     Arcene & 0.13 $\pm$ 0.004 & 0.79 $\pm$ 0.874\\
     \hline
     Compas &  0.99 $\pm$ 0.009 & 1.23 $\pm$ 1.758 \\
     \hline
     Dexter &  0.217 $\pm$ 0.005 & 2.76 $\pm$ 2.452\\
     \hline
     German & 0.11 $\pm$ 0.004 & 1.3 $\pm$ 0.677 \\
     \hline
    \end{tabular}
\end{table}
\vspace{-.6cm}

Table \ref{tab:tt_fptc_rf} reports the mean and standard deviation of the actual training time and the predicted one for the Random Forest classifier. Differently from above, in this case, we notice a higher variability among the predicted training times, especially in \textit{Adult}, \textit{APS}, \textit{Compas}, and \textit{Dexter}. In addition, we notice how for the \textit{APS} dataset (which is the one letting the worse performances), the FPTC method underestimates the real training time. Finally, as noticed above, the low training time of some datasets (namely, \textit{Antivirus}, \textit{Arcene}, \textit{Dexter}) explains the high value of the related MAPE metric for them. 

From this analysis, we can conclude how the FPTC method is able to predict the training time of a Logistic Regression and Random Forest classifier under certain circumstances (\ie datasets) while it is not working in others. However, we do not notice any correlation between specific characteristics of the dataset (\eg number of features) and the correctness of the predictions. Moreover, we see how the correctness of the predictions is directly correlated to the value of the slope, which is again not only dependent on the execution environment but also varies with the variation of the dataset used to compute it, as shown in Section \ref{sec:exp_one}.

\begin{shadedbox}
    \textbf{Answer to \ref{rq2}:} The FPTC method is able to predict the training time of the Logistic Regression and Random Forest classifiers under certain circumstances (\ie datasets), while it fails in others. The correctness of the predictions (especially for the Random Forest classifier) is strongly related to the value of the slope, which depends on the dataset used to compute it. 
\end{shadedbox} 

\subsection{Threats to Validity}

\noindent\textbf{Internal validity:} We adopted a synthetic dataset to compute the slopes to answer \ref{rq1}. In contrast, a real-world dataset could include more complexity and variability not considered in this experiment. To answer this threat, we clarify that the goal of our experiment was to prove that the value of the slope is not only dependent on the execution environment. Hence, any dataset (synthetic or not) that proves this hypothesis is effective. 

\noindent\textbf{External validity:} The results of our experiments may apply only to the selected ML models and datasets. Concerning the selection of the dataset, we selected several datasets heterogeneous in their dimensions, making our results enough general. Concerning the ML models, we analysed two of the most adopted ML models for classification, while we will analyse the others in future works.

\vspace{-.3
cm}
\section{Conclusion and Future Work}\label{sec:conclusion}
\vspace{-.3cm}

In this paper, we have presented the work we are conducting towards predicting the training time of ML models. In particular, we have extensively evaluated the work proposed in \cite{zheng2021full}, which is the only approach so far that formulates the training time as a function of the dataset's and model's parameters. In this paper, we have considered the formulations proposed for the Logistic Regression and Random Forest classifiers, and we have shown how the proposed approach is not always able to predict the training time successfully. Further, from the results shown in Section \ref{sec:rq2}, there is no evidence of any correlation between the dataset size and the correctness of the predictions. 
Instead, from the results shown in Section \ref{sec:exp_one}, there is a correlation between the number of dataset features and the value of the slope used in the FPTC formulation (which is, again, not only dependent on the execution environment as stated in \cite{zheng2021full}).

In the future, we want to deeper analyse the formulations proposed for the different ML models and overcome the observed limitations. In particular, we want to investigate if some specific characteristics of the dataset or ML model influence the training time and are not considered in the current formulation. 

\bibliographystyle{splncs04}
\bibliography{bibliography}
\end{document}